\documentclass[]{OAGM}
\OAGMarXiv{1505.01065}

\usepackage[utf8]{inputenc}
\usepackage[english]{babel}
\usepackage{amsmath}
\usepackage{amsfonts}
\usepackage{amssymb}
\usepackage{graphicx}
\usepackage{caption}
\usepackage{subcaption}

\title{Using Dimension Reduction to Improve the Classification of High-dimensional Data}

\author{Andreas Gr\"{u}nauer and Markus Vincze\\
    Automation and Control Institute, Vienna University of Technology, Austria\\
    {\tt\footnotesize \{gruenauer, vincze\}@acin.tuwien.ac.at} } 

\begin{document}
\maketitle

\begin{abstract}
In this work we show that the classification performance of high-dimensional structural MRI data with only a small set of training examples is improved by the usage of dimension reduction methods. We assessed two different dimension reduction variants: feature selection by ANOVA F-test and feature transformation by PCA. On the reduced datasets, we applied common learning algorithms using 5-fold cross-validation. Training, tuning of the hyperparameters, as well as the performance evaluation of the classifiers was conducted using two different performance measures: Accuracy, and Receiver Operating Characteristic curve (AUC). Our hypothesis is supported by experimental results.
\end{abstract}

\section{Introduction}

Machine learning algorithms are used in various fields for learning patterns from data and make predictions on unseen data. Unfortunately, the probability of overfitting of a learning algorithm increases with the number of features~\cite{Verleysen_curse}. Dimension reduction methods are not only powerful tools to avoid overfitting~\cite{ng_feature_selection}, but also capable of making the training of high-dimensional data a computationally more feasible task. In this work, we want to study the influence of dimension reduction techniques on the performance of various well-known classification methods. Dimension reduction methods are categorized into two groups: feature selection and feature transformation. 

Feature selection methods~\cite{mladenic_feature_selection,Guyon_intro_fs} aim to identify a subset of ``meaning-ful'' features out of the original set of features. They can be subdivided into filter, wrapper and embedded methods. Filter methods compute a score for each feature, and then select only the features that have the best scores. Wrapper methods train a predictive model on subsets of features, before the subset with the best score is selected. The search for subsets can be done either in a deterministic (e.g. forward selection, backward elimination) or random (i.e. genetic algorithms) way. Embedded methods determine the optimal subset of features directly by the trained weights of the classification method. 

In contrast to feature selection methods, feature transformation methods project the original high dimensional data into a lower dimensional space. Principal Component Analysis (PCA) is one of the most-known techniques in this category. PCA finds the principal axis in the dataset that explain most of the variance, without considering the class labels. Therefore we use PCA as the baseline for dimension reduction methods in this study.

Among various feature selection methods, we limit our scope on filter methods, as they do not depend on a specific classification method and therefore are suitable for the comparison of different classifiers~\cite{mladenic_feature_selection}. Comparative studies on text classification have revealed, that univariate statistical tests like the $\chi^2$-test and ANOVA F-test are among the most effective scores for filtered feature selection~\cite{yang_fs_study}. As the $\chi^2$-test is only applicable on categorical data, we use the filter selection method based on the ANOVA F-test which is applicable on continuous features used for the evaluation of this study.

As part of the 17th International Conference on Medical Image Computing and Computer Assisted Intervention (MICCAI), the MICCAI 2014 Machine Learning Challenge (MLC) aims for the objective comparison of the latest machine learning algorithms applied on Structural MRI data~\cite{MLC2014}. The subtask of binary classification of clinical phenotypes is in particular challenging, since in the opinion of the challenge authors, a prediction accuracy of 0.6 is acceptable. Motivated by this challenge, the goal of this study is to show, that the selected dimension reduction methods improve the performances of various classifiers trained on a small set of high-dimensional Structural MRI data.

The report is organized as follows. Section~\ref{sec:Dimension reduction} describes the dimension reduction methods. Section~\ref{sec:Classifiers} describes the classifiers. Section \ref{sec:Experiment settings} describes the datasets, evaluation measures and the experiment methodology. Section~\ref{sec:Results} presents the results. Section~\ref{sec:Discussion} discusses the major findings. Section~\ref{sec:Conclusion} summarizes the conclusion.

\section{Dimension reduction}
\label{sec:Dimension reduction}
In the following we will give an overview of the used techniques for dimension reduction.

\paragraph{Filtered feature selection by ANOVA F-test}
Feature selection methods based on filtering determine the relevance of features by calculating a score (usually based on a statistical measure or test). Given a number of selected features $s$, only the $s$ top-scoring features are afterwards forwarded to the classification algorithm. In this study, we use the ANOVA F-Test statistic~\cite{statistical_concepts} for the feature scoring. The F-test score assesses, if the expected values of a quantitative random variable $x$ within a number of $m$ predefined groups differ from each other. The F-value is defined as
\[ F=\frac{{MS}_B}{{MS}_W},\]
${MS}_B$ reflects the ``between-group variability'', expressed as
\[ {MS}_B= \frac{\sum_i n_i(\bar{x}_i - \bar{x})^2 }{m-1}, \]
where $n_i$ is the number of observations in the $i$-th group, $\bar{x}_i$ denotes the sample mean in the $i$-th group, and $\bar{x}$ denotes the overall mean of the data.
${MS}_W$ refers to the ``within-group variability', defined as
\[ {MS}_W= \frac{\sum_{ij} (x_{ij} - \bar{x_i})^2 }{n-m}, \]
where $x_{ij}$ denotes the $j$-th observation in the $i$-th group. For the binary classification problem assessed in this report, the number of groups $m=2$.

\paragraph{Feature transformation by PCA}
PCA~\cite{jolliffe_pca} reduces the dimension of the data by finding the first $s$ orthogonal linear combinations of the original variables with the largest variance. PCA is defined in such a way that the first principal component has the largest possible variance. Each succeeding component in turn has the highest variance possible under the constraint that it is orthogonal to the preceding components.

\section{Classifiers}
\label{sec:Classifiers}

In the following we give a short description of each used classification methods in this study.

\paragraph{k-Nearest Neighbors (k-NN)}

The k-NN classifier~\cite[p.125]{bishop} does not train a specific model, but stores a labeled training set as ``reference set''. The classification of a sample is then determined by the class with the most representatives among the $k$ nearest neighbors of the sample in the reference set. Odd values of $k$ prevent tie votes. Among other possible metrics, we use the Euclidean distance metric for this study.

\paragraph{Gaussian Naive Bayes (GNB)}

Bayes classifiers are based on the Bayes' theorem and depend on naive (strong) independence assumptions~\cite{zhang_gaussianNB,duda2001pattern}. Using Bayes’ theorem, the probability $P(\omega_j \mid \mathbf{x})$ of some class $\omega_w$ given a $d$-dimensional random feature vector $\mathbf{x} \in \mathbb{R}^d$ can be expressed as the equation:

\[P(\omega_j \mid \mathbf{x}) = \frac{P(\omega_j) p(\mathbf{x} \mid \omega_j)}{p(\mathbf{x})},\]
where $P(\omega_j)$ denotes the prior probability of the $j$-th class $\omega_j$, $p(\mathbf{x} \mid \omega_j)$ refers to the class conditional probability density function for $\mathbf{x}$ given class $\omega_j$, and $p(\mathbf{x})$ is the evidence factor used for scaling, which in the case of two classes is defined as
\[ p(\mathbf{x}) = \sum_{j=2}^2p(\mathbf{x} \mid \omega_j)P(\omega_j).\]
Under the \textit{naive} assumption that all the individual components $x_i, i=1, \dots, d$ of $\mathbf{x}$ are conditionally independent given the class, $p(\mathbf{x} \mid \omega_j)$ can be decomposed into the product $p(x_1 \mid \omega_j) \dots p(x_d \mid \omega_j)$. Therefore we can rearrange $P(\omega_j \mid \mathbf{x})$ as

\[ P(\omega_j \mid \mathbf{x}) = \frac{P(\omega_j)\prod_{i=1}^{d} p(x_i \mid \omega_j)}{p(\mathbf{x})}. \]
Since $p(\mathbf{x})$ is constant for a given input under the Bayes' rule, the \textit{naive Bayes classifier} predicts the class $\omega_k$ that maximizes the following function:
\[ \omega_k = \arg\max_{j} P(\omega_j)\prod_{i=1}^{d} p(x_i \mid \omega_j) \]
Under the typical assumption that continuous values associated with each class are Gaussian distributed, the probability density of a component $x_i$ given a class $\omega_j$ can be expressed as
\begin{align*}
p(x_i \mid \omega_j) &= \frac{1}{\sqrt{2\pi\sigma^2_{ij}}} \exp\left(-\frac{ (x_i - \mu_{ij})^2}{2\pi\sigma^2_{ij}}\right),
\end{align*}
where $\mu_{ij}$ denotes the class conditional mean and $\sigma^2_{ij}$ the class conditional variance.
The corresponding classifier is called \textit{Gaussian Naive Bayes}.

\paragraph{Linear Discriminant Analysis (LDA)}

Given a two class problem, the LDA algorithm separates the projected mean of two classes maximally by a defined separating hyperplane, while minimizing the variance within each class~\cite[p. 117--124]{duda2001pattern}. LDA is based on the assumption that both classes are normally distributed and share the same covariance matrix.

\paragraph{Ridge}

The Ridge classifier is based on Ridge Regression, which extends the Ordinary Least Squares (OLS) method with an additional penalty term to limit the $L^2$-norm of the weight vector~\cite{rifkin_ridge}. This penalty term shrinks the weights to prevent overfitting.


\paragraph{Support Vector Machine (SVM)}
A support vector machine~\cite[p. 325]{bishop} solves the classification of a dataset by constructing a hyperplane in a high or infinite dimensional space in such a way that the distance of the nearest points of the training data to the hyperplane is maximized. The idea of the large margin is to ensure, that samples which are not exactly equal to the training data can still be classified in a reliable way. To prevent overfitting by permitting some degree of misclassifications, a cost parameter $C$ controls the trade off between allowing training errors and forcing rigid margins. Increasing the value of $C$ increases the cost of misclassifying points and forces the creation of a more accurate model that may not generalize well. 
For our experiment we use a SVM classifier with linear kernel (SVM-L) as well as a SVM classifier with a non-linear kernel using radial basis functions (SVM-RBF).

\paragraph{Random Forests (RF)}
Bagging predictors ~\cite{breiman_randomforests} generate multiple versions of a predictor (in this case decision trees) which are used to get an aggregated predictor. By generating a set of trees in randomly selected subspaces of the feature space~\cite{amit_rforests}, the different trees generalize their classification in complementary ways.

\section{Experiment settings}
\label{sec:Experiment settings}

In the following section we illustrate the conducted experiments in detail. In this study we used the machine learning library scikit-learn~\footnote{\url{http://scikit-learn.org}} version 0.14.1 for all proposed methods and scoring measures in this study. This open source Python library provides a wide variety of machine learning algorithms based on a consistent interface, which eases the comparison of different methods for a given task.

\subsection{Dataset}
In our experiments, we used the dataset for the binary classification task of the MLC 2014~\cite{MLC2014}. This dataset consists of 250 T1-weighted structural brain MRI scans: 150 scans including the target class labels for training and additional 100 samples with unknown class labels, reserved for the challenge submission. For each scan, a number of 184 morphological summary features are provided. These features represent volumes of cortical and sub-cortical structures, as well as average thickness measurements within cortical regions. The volume measures have been normalized with the intracranial volume (ICV) to account for different head sizes. All features have been extracted using the brain MRI software \textit{FreeSurfer}~\cite{fischl_freesurfer}.

\subsection{Evaluation measures}

As recommended by the MLC 2014 challenge~\cite{MLC2014}, we used two common performance measures: Accuracy and the area under the Receiver Operating Characteristic curve (AUC). Both compare the predictions of the classifier with the groundtruth provided in the training data. 
\paragraph{Accuracy}
The accuracy is defined as follows:
\[accuracy = \frac{tp+tn}{tp+fp+tn+fn},\]
where where $tp$, $tn$, $fp$, $fn$ present the number of true positives, true negatives, false positives and false negatives, respectively.
\paragraph{Area under the ROC curve (AUC)}
The ROC curve presents the tradeoff between the true positive rate (TPR), expressed as 
\[ \mathit{TPR} =  \frac{\mathit{tp}}{\mathit{tp}+\mathit{fn}}, \]

and the false positive rate (FPR), defined as
\[ \mathit{FPR} =  \frac{\mathit{fp}}{\mathit{fp}+\mathit{tn}}. \] Given a two class problem, a ROC curve can be plotted by varying the probability threshold for predicting positive examples in an interval between zero and one. Informally, one point in ROC space is better than another if it is to the northwest ($tp$ rate is higher, $fp$ rate is lower, or both) of the first. Hence the curve visualizes, for what region a model is more superior compared to another. The area under the ROC curve (AUC) maps this relation to a single value.

\subsection{5-Fold cross-validation (CV)}
We used 5-fold CV by randomly splitting the training dataset ($D$) of 150 samples into five mutually exclusive subsets ($D_1, D_2, D_3, D_4, D_5$) of approximately equal size. Each classification model was trained and tested five times, where each time ($t \in \{1, 2, 3, 4, 5\}$), it was trained on all except one fold ($D \setminus D_t$) and tested on the remaining fold ($D_t$). The accuracy and AUC measures were averaged over the particular measures of the five individual test folds.

\subsection{Experiment Methodology}

Our experiments were conducted in the following way. We applied each dimension reduction method on the original training set with a different number of $s \in\{3,6,12,24,48,92,184\}$ selected components. We trained the classifiers on the 150 datasets with known target class labels using 5-fold CV in two ways: the first by optimizing the accuracy measure and the second by optimizing the AUC measure. For classifiers based on a set of specific hyperparameters, we performed a grid search to find the optimal configuration of hyperparameters. As an exhaustive search over all possible hyperparameters would be an unfeasible task, we limited our scope on a subset of hyperparameters for each classifier with a discrete set of tested values. Table~\ref{table:hyperparameters} shows the selected hyperparameters and the corresponding set of values for each classifier.

\begin{table}[b]
\centering
\caption{The selected hyperparameters and corresponding values for hyperparameter optimization using grid search.}
	\begin{tabular}{|l|l|}
	\hline
	\textbf{Classifier} & \textbf{Tuned hyperparameters}                                                                           \\
	KNN        & $k \in \{3,5,7,9,11,13,15\}$                                                               \\
	Ridge      & $\alpha \in \{0.1, 1, 10\}$\\
	SVM        & $C \in \{10^0,10^1, \dots, 10^8\}$\\                                                                       
	SVM-RBF & $\gamma \in \{10^{-10},10^{-9}, \dots, 10^2\}$\\
					&  $C \in \{10^0,10^1, \dots, 10^8\}$\\
	RF         & $N_{trees} \in \{2,4,8,16,32\},$                                                       \\                                                                                          
				& with number of trees $N_{trees}$\\
	\hline
	\end{tabular}
\label{table:hyperparameters}%
\end{table}

\section{Results}
\label{sec:Results}

        

\begin{figure}[t]
        \centering
        \begin{subfigure}[b]{0.43\textwidth}
                \includegraphics[width=\textwidth]{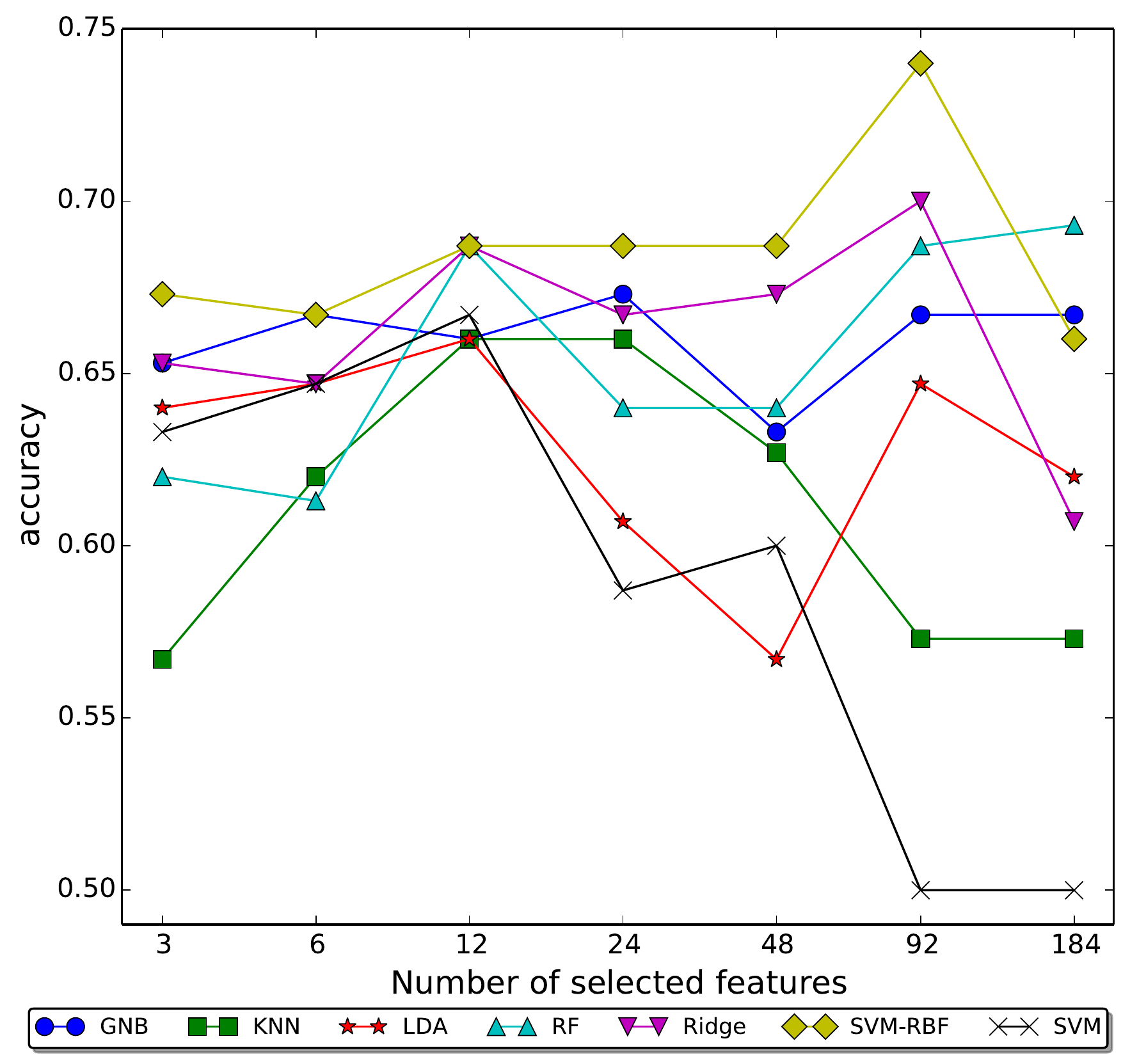}
                \caption{Hyperparameter tuning and evaluation using accuracy}
                \label{fig:uvanova.accuracy}
        \end{subfigure}%
        ~ 
        \begin{subfigure}[b]{0.43\textwidth}
                \includegraphics[width=\textwidth]{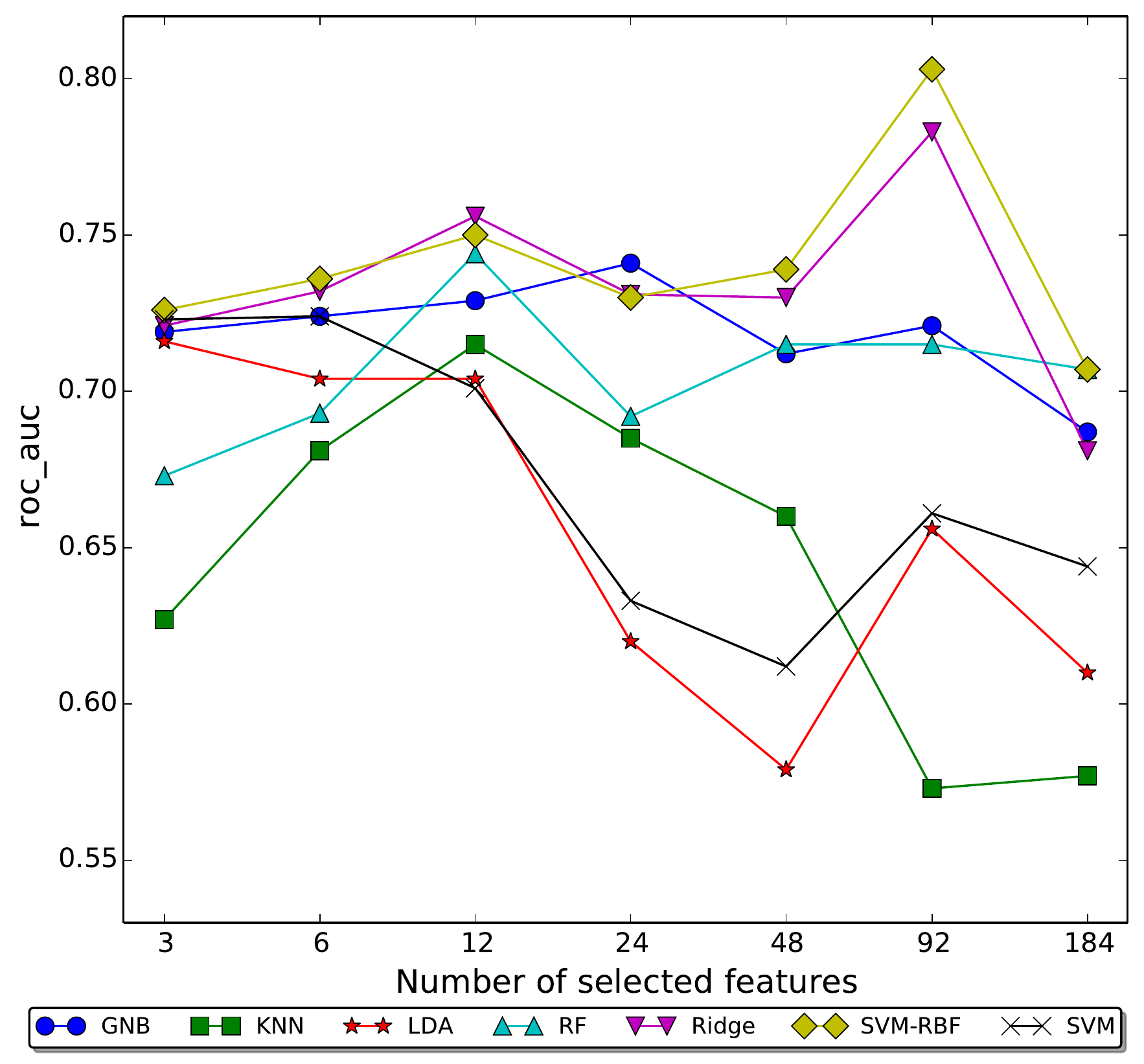}
                \caption{Hyperparameter tuning and evaluation using AUC}
                \label{fig:uvanova.roc_auc}
        \end{subfigure}
        \caption{Classifier performances using $\mathit{ANOVA}$-based feature selection.}\label{fig:uvanova}
\end{figure}

\begin{figure}[t]
        \centering
        \begin{subfigure}[b]{0.43\textwidth}
                \includegraphics[width=\textwidth]{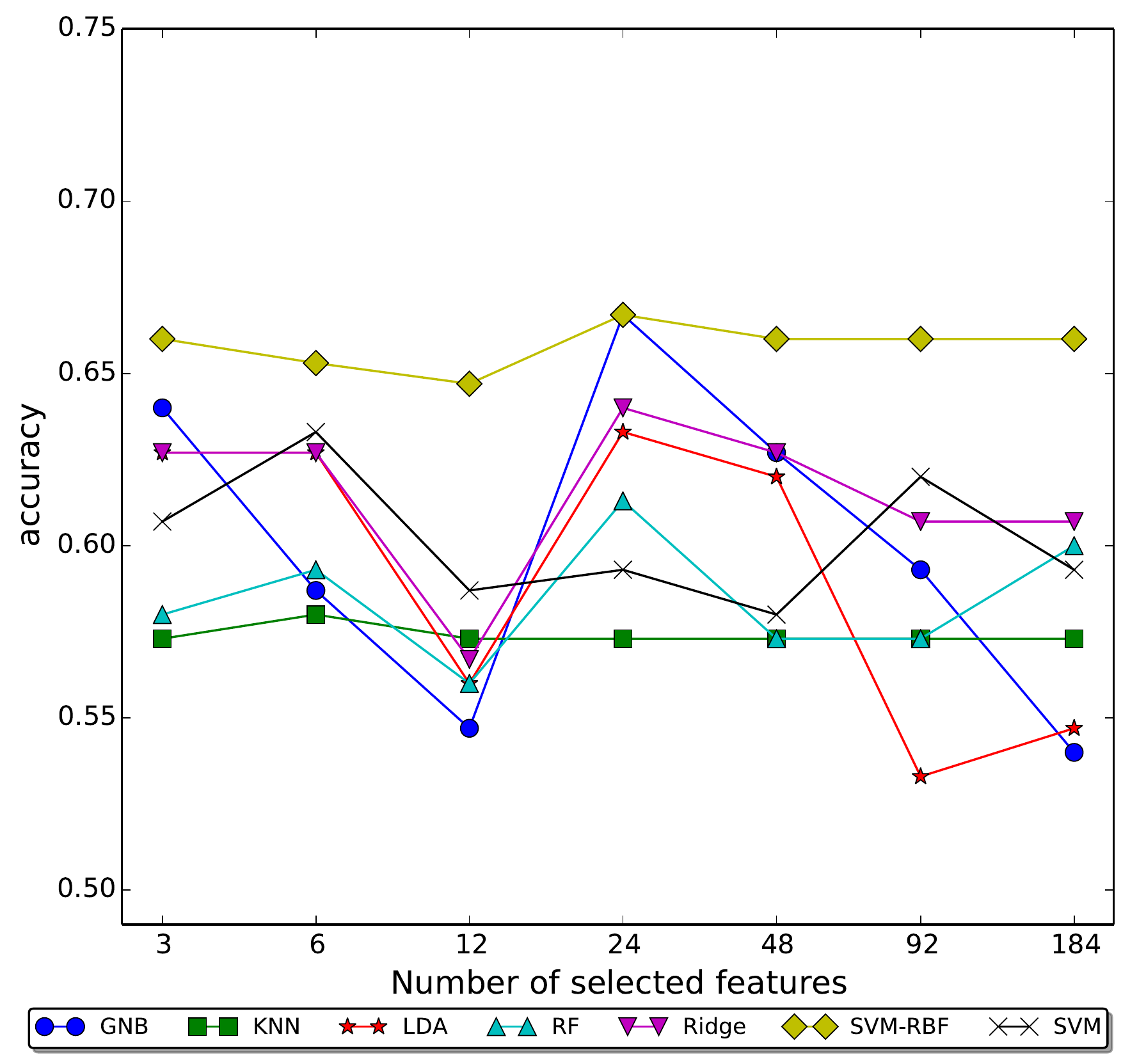}
                \caption{Hyperparameter tuning and evaluation using accuracy}
                \label{fig:pca.accuracy}
        \end{subfigure}%
        ~ 
        \begin{subfigure}[b]{0.43\textwidth}
                \includegraphics[width=\textwidth]{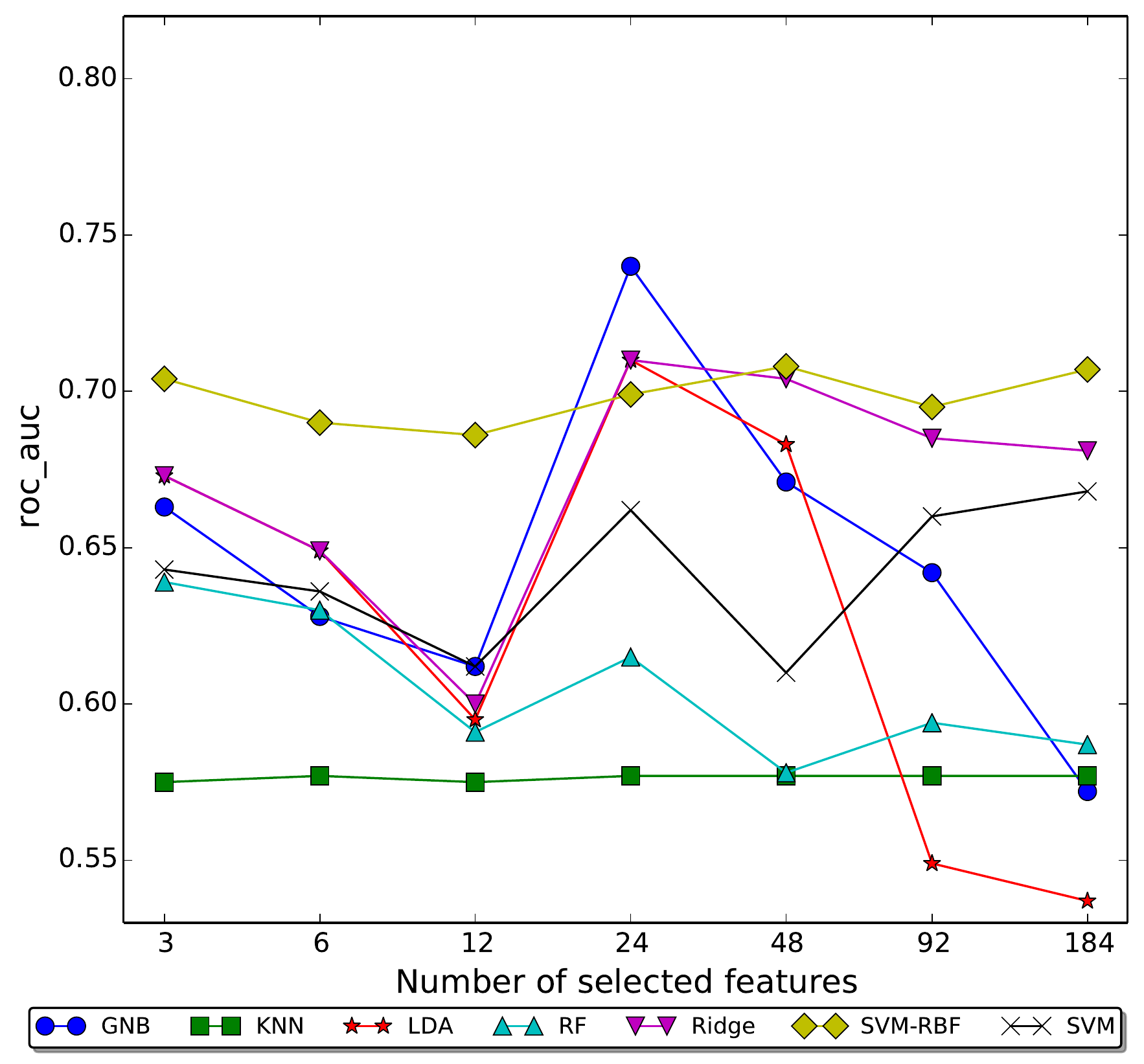}
                \caption{Hyperparameter tuning and evaluation using AUC}
                \label{fig:pca.roc_auc}
        \end{subfigure}
        \caption{Classifier performances using PCA-based dimension reduction.}\label{fig:pca}
\end{figure}


Fig.~\ref{fig:uvanova} shows the performance of the classifiers on the basis of ANOVA F-test feature selection using accuracy (Fig.~\ref{fig:uvanova.accuracy}) and AUC (Fig.~\ref{fig:uvanova.roc_auc}) for hyperparameter-tuning and performance evaluation, respectively. Both figures reveal that, with $s=12$ selected features, the classifiers achieve already the same or better performances than using the original $s=184$ features.
When the number of selected features is further increased, the performance of the RBF-SVM peaks at $s=92$, while the performances of the other classifiers do not improve or rather decline. 
This observation shows the importance of feature selection, as more features do not necessarily lead to better performance (overfitting).

Fig.~\ref{fig:pca} displays the classifier performance on the basis of PCA-reduced data using accuracy (Fig.~\ref{fig:pca.accuracy}) and AUC (Fig.~\ref{fig:pca.roc_auc}) for hyperparameter-tuning and performance evaluation, respectively. Both figures show that the performances of SVM-RBF and KNN are both independent from the amount of used components, with the difference that SVM-RBF outperforms the other classifiers, while KNN exhibits a constantly weak performance over all used components.
The other classifiers perform already better on the first $s=3$ components of the PCA, than on the original features. When the number of used components $s$ is further increased, the classifiers show a common performance breakdown at $s=12$. Increasing the number of components $s$ leads to the performance maximum at a number of $s=24$ components.

In this study we additionally observe that the GNB classifier performs better than the LDA classifier for a number of $s > 12$ selected components, although both methods share the assumption that the random variables are independent from each other and normally distributed. The key difference is that the LDA method additionally considers the covariance of the dimensions. When the number of samples is lower than the number of dimensions, like it is the case in this study, the accurate estimation of the covariance matrix can not be guaranteed. This phenomenon is known as the ``small sample size'' problem~\cite{raudys_small_sample}. This observation suggets that, due to its simpler assumptions, GNB is a more robust classification method than LDA, given a small sized training set.

\section{Discussion}
\label{sec:Discussion}
The performances of the majority of investigated classifiers converge consistently at the same number of $s$ selected features, independent of the measure used for the tuning of hyperparameters. This indicates that the search for the optimal number of selected features is a robust way to improve the performance of classifiers given high dimensional data. The results confirm that the RBF-SVM classifier outperforms the other classifiers independent from the number of reduced features. But the results also show that linear classifiers like GNB and Ridge are able to produce equal or even better results on reduced dimensions using the chosen feature selection methods than the RBF\hbox{-}SVM classifier.

\section{Conclusion}
\label{sec:Conclusion}
The performances of classifiers under various scores for hyperparameter tuning combined with different dimension-reduction methods are analyzed.
Both dimension reductions improved the performance of all classifiers in comparison to the original high-dimensional data. The results indicated that ANOVA F-Test feature selection yielded the better results compared to the PCA-based feature transformation.

\subsection*{Acknowledgments}
\small
The research leading to these results has received funding from the European Community, Horizon 2020 Programme (H2020-ICT-2014-1), under grant agreement No. 641474, FLOBOT. This work was supported by the Pattern Recognition and Image Processing Group at TU Wien. I thank Yll Haximusa and Roxane Licandro for their thoughtful comments and proofreading.

\bibliography{references}
\end{document}